\documentclass[conference,a4paper]{IEEEtran}
\IEEEoverridecommandlockouts

\usepackage[hidelinks]{hyperref}
\usepackage[cmex10]{amsmath}
\usepackage{amssymb,amsfonts}
\usepackage{dblfloatfix}

\usepackage[ruled,vlined]{algorithm2e}
\usepackage{graphicx}
\graphicspath{{Figures/PDF/}{Figures/PNG/}}

\usepackage{booktabs}
\usepackage{siunitx}
\usepackage[numbers,compress]{natbib}
\usepackage{bm,bbm}
\usepackage{orcidlink}
\usepackage{multirow} % for table headers (optional, but safe to keep)

\begin{document}

\title{\uppercase{Implicit Spatial-Frequency Fusion of Hyperspectral and LiDAR Data via Kolmogorov-Arnold Networks}}

% \author{
% \IEEEauthorblockN{Zekun Long}
% \IEEEauthorblockA{\textit{Griffith University}\\
% Brisbane, QLD, Australia\\
% (e-mail: zekun.long@griffithuni.edu.au)}
% \and
% \IEEEauthorblockN{Judy X.~Yang}
% \IEEEauthorblockA{\textit{Griffith University}\\
% Brisbane, QLD, Australia\\
% (e-mail: judy.yang@griffithuni.edu.au)}
% \and
% \IEEEauthorblockN{Jing Wang}
% \IEEEauthorblockA{\textit{Griffith University}\\
% Brisbane, QLD, Australia\\
% (e-mail: jing.wang@griffith.edu.au)}
% \and
% \IEEEauthorblockN{Ali Zia}
% \IEEEauthorblockA{\textit{La Trobe University}\\
% Melbourne, VIC 3086, Australia\\
% (e-mail: a.zia@latrobe.edu.au)}
% \and
% \IEEEauthorblockN{Guanyiman Fu}
% \IEEEauthorblockA{\textit{Griffith University}\\
% Brisbane, QLD, Australia\\
% (e-mail: guanyiman.fu@griffith.edu.au)}
% \and
% \IEEEauthorblockN{Jun Zhou}
% \IEEEauthorblockA{\textit{Griffith University}\\
% Brisbane, QLD, Australia\\
% (e-mail: jun.zhou@griffith.edu.au)}
% }
\author{
\IEEEauthorblockN{Zekun Long\IEEEauthorrefmark{1},
Judy X.~Yang\IEEEauthorrefmark{1},
Jing Wang\IEEEauthorrefmark{1},
Ali Zia\IEEEauthorrefmark{2},
Guanyiman Fu\IEEEauthorrefmark{1},
Jun Zhou\IEEEauthorrefmark{1}}
\IEEEauthorblockA{\IEEEauthorrefmark{1}
School of Information and Communication Technology,
Griffith University, Nathan, Australia}
\IEEEauthorblockA{\IEEEauthorrefmark{2}
School of Computing, Engineering and Mathematical Sciences, La Trobe University, Melbourne, Australia}
}

% \author{
% Zekun~Long,
% Judy~X.~Yang,
% Jing~Wang,
% Guanyiman~Fu,
% Ali~Zia,
% and Jun~Zhou
% \thanks{
% Z. Long, J. X. Yang, J. Wang, G. Fu, and J. Zhou are with the School of Information and Communication Technology,
% Griffith University, Brisbane, QLD 4111, Australia.
% }
% \thanks{
% A. Zia is with the Department of Computer Science and Information Technology,
% La Trobe University, Melbourne, VIC 3086, Australia.
% }
% \thanks{
% Zekun Long (e-mail: zekun.long@griffithuni.edu.au).
% }
% \thanks{
% Corresponding author: Jun Zhou, Fellow, IEEE (e-mail: jun.zhou@griffith.edu.au).
% }
% }

\maketitle

\begin{abstract}
% Hyperspectral image (HSI) classification benefits from rich spectral signatures but often suffers from spectral ambiguity and limited spatial context, especially in complex urban scenes. LiDAR provides complementary geometric cues, yet existing HSI-LiDAR fusion methods typically rely on predefined fusion heuristics or heavy attention-based designs on discrete grids, which limits their ability to model spatial-spectral-geometric interactions efficiently. In this paper, we propose the \emph{Implicit Frequency-Geometry Fusion Network} (IFGNet), which formulates multimodal fusion as learning geometry-conditioned implicit aggregation functions parameterized by Kolmogorov-Arnold Networks (KANs). IFGNet performs implicit aggregation in two domains: in the spatial domain, LiDAR-guided geometric information reshapes local representations; in the frequency domain, Fourier-domain modeling captures global structural patterns and long-range dependencies. Extensive experiments on the Houston 2013 and MUUFL benchmarks demonstrate that IFGNet consistently achieves state-of-the-art performance in terms of OA, AA, and Cohen's Kappa, while maintaining a lightweight architecture suitable for practical deployment.
Hyperspectral image (HSI) classification is challenging in complex scenes due to spectral ambiguity, spatial heterogeneity, and the strong coupling between material properties and geometric structures. Although LiDAR provides complementary elevation information, most HSI-LiDAR fusion methods rely on CNN or MLP with fixed activation functions and linear weights. They struggle to model structural discontinuities of LiDAR, intricate spectral features of HSI, and their interactions. In addition, fusing the two data modalities in both spatial and frequency domains with LiDAR guidance is not fully explored in the literature. To address these issues, we propose the \emph{Implicit Frequency-Geometry Fusion Network} (IFGNet), which leverages Kolmogorov-Arnold Networks (KANs) with learnable spline-based functions to adaptively capture highly non-linear relationships and enable adaptive integration of hyperspectral and LiDAR features. Furthermore, IFGNet introduces a LiDAR-guided implicit aggregation module in spatial and frequency domains, which enhances geometry-aware spatial representations, and captures global structural patterns. Experiments on the Houston 2013 and MUUFL benchmarks demonstrate that IFGNet consistently outperforms existing fusion methods in overall accuracy, average accuracy, and Cohen’s Kappa, while maintaining an efficient architecture.

\end{abstract}


\section{Introduction}

Hyperspectral image (HSI) classification plays a critical role in urban mapping~\cite{Benediktssonurban}, environmental monitoring~\cite{Alipourfard}, and land-cover analysis~\cite{vali2020deep}. By capturing hundreds of contiguous spectral bands, HSI provides fine-grained spectral signatures that enable material-level discrimination~\cite{amigo2015hyperspectral}. However, high spectral dimensionality, spectral redundancy, and sensitivity to illumination variation pose significant challenges to robust classification~\cite{li2019deep}. LiDAR data, which encodes elevation and structural information, has been widely adopted as a complementary modality to mitigate spectral ambiguity, particularly in complex urban scenes where objects may share similar spectral responses, but differ in height or geometry~\cite{zhao2025mgf}.

Most HSI-LiDAR fusion approaches have relied on explicit modeling strategies or joint representations based on deep learning~\cite{Ghamisifusion,Hongfusion}. Traditional explicit fusion methods typically combine hyperspectral spectral information with LiDAR-derived geometric features through dimensionality reduction techniques and handcrafted feature design~\cite{Jahanfusion}. Although these methods exploit cross-modal complementarity to some extent, their performance heavily depends on manually designed features and prior assumptions, limiting adaptability to complex scenes and nonlinear relationships~\cite{audebert2019deep}. 

% Moreover, direct concatenation of high-dimensional features often leads to the curse of dimensionality, increasing training difficulty and degrading performance~\cite{Chhapariyafusion}.

Deep-learning-based methods have further advanced HSI-LiDAR fusion by enabling end-to-end feature learning~\cite{RoyFusionTransformer, wang2021spectral}. Most works adopt dual-branch convolutional neural networks (CNNs) or transformer to separately extract features from HSI and LiDAR data, followed by feature fusion at intermediate or high-level representations~\cite{zhang2025adaptive,yang2025hslinets}. Due to the limitation of local convolutional kernels, transformer-based fusion methods have become more popular for their capability to model long-range dependencies.

% While these approaches achieve notable improvements in classification accuracy, CNNs primarily rely on local convolutional kernels, which limits their capability to model fine-grained spectral variations and long-range spectral-spatial dependencies inherent in hyperspectral data~\cite{rehman2025deep}.  

% However, conventional CNNs rely on fixed local receptive fields and exhibit limited nonlinear representation capacity, while Transformer-based models incur substantial computational overhead. 
Despite these advances, existing HSI and LiDAR fusion methods suffer from the following issues: first, the basic units of CNN and transformer are based on fixed activations and linear weights multiplication, which limits the representation capability for the complex HSI and LiDAR data in the fusion. Second, the current HSI and LiDAR fusion strategy predominantly extracts and merges features of the two modalities as two branches. These methodologies lack LiDAR-guided aggregation fusion, as they often treat structural and spectral data as independent inputs rather than leveraging the geometric precision of LiDAR to actively steer the sampling and refinement of spectral context. Third, fusing HSI and LiDAR from spatial and frequency domains has proven its preliminary value, but has not been fully explored.

Recently, Kolmogorov-Arnold Networks (KANs)~\cite{liu2024kan} have been proposed as a powerful alternative to the traditional unit in neural networks.  KANs are a class of neural networks inspired by the Kolmogorov-Arnold representation theorem. Unlike traditional neural networks that use linear transformations followed by fixed nonlinear activations, KANs parameterize nonlinear mappings using learnable spline functions, enabling more flexible and efficient function approximation with fewer parameters. It has shown successful applications in computer vision and time series analysis~\cite{somvanshi2025survey}.

To address these issues and inspired by KAN, we propose the \emph{Implicit Frequency-Geometry Fusion Network} (IFGNet) for HSI-LiDAR classification. Instead of performing fusion through explicit feature concatenation or discrete attention, IFGNet models cross-modal interactions via implicit aggregation functions parameterized by Kolmogorov-Arnold Networks (KANs)~\cite{liu2024kan}. In the proposed framework, hyperspectral and LiDAR features are encoded into a shared latent space and implicitly aggregated across spatial and frequency domains for patch-based classification. By decoupling spatial-geometric aggregation from frequency domain modeling and unifying them under an implicit representation fusion framework, IFGNet achieves effective multimodal fusion while maintaining a lightweight architecture. This design also avoids heavy attention modules and deep transformer networks, making it suitable for practical remote sensing scenarios. 

The main contributions of this work are summarized as follows:
\begin{itemize}
\item \textbf{KAN-Driven Nonlinear Feature Modeling:} We pioneer the use of Kolmogorov-Arnold Networks (KANs) for HSI-LiDAR fusion, replacing rigid linear weights with learnable spline-based functions. This enables the model to parameterize spatial and frequency fusion operators that adaptively capture complex, nonlinear cross-modal interactions conditioned on both spectral content and continuous spatial coordinates.
\item \textbf{LiDAR-Guided Spatial Implicit Aggregation:} We introduce a guided aggregation mechanism that utilizes LiDAR-derived geometric cues as a structural reference. By sampling neighboring hyperspectral features through an implicit neural representation, the model achieves adaptive, area-aware fusion that prioritizes structurally consistent regions and preserves sharp elevation boundaries often blurred by traditional convolution-based methods.
\item \textbf{Dual-Domain Implicit Fusion Framework:} Our method models both spatial-domain geometric aggregation and frequency-domain structural dependencies. By performing implicit fusion across both domains, the network avoids predefined heuristics and captures a more comprehensive representation of the high-frequency phase information and global spectral consistency inherent in multi-modal remote sensing data.
% \item \textbf{KAN-driven feature extraction and aggregation: }
% Kolmogorov-Arnold Networks are employed to extract HSI and LiDAR features, and parameterize spatial and frequency fusion operators, enabling flexible nonlinear cross-modal interactions conditioned on feature content and spatial coordinates.

% \item \textbf{LiDAR-guided spatial area fusion: }
% LiDAR-derived geometric cues are exploited to guide spatial aggregation of hyperspectral features, enabling adaptive area-aware fusion that emphasises structurally consistent regions and suppresses irrelevant spatial responses.

% \item \textbf{Implicit frequency-geometry fusion: }
% We propose IFGNet, a novel HSI-LiDAR fusion network that jointly models spatial-domain geometric aggregation and frequency-domain spectral structures through implicit representations, avoiding predefined fusion heuristics.  
\end{itemize}

\section{Methodology}
\label{sec:method}

\subsection{Overall Framework}
Let $\mathbf{X}_{p}\in\mathbb{R}^{P^2\times T}$ denote an HSI patch centered at a labeled pixel, where $P$ is the spatial patch size and $T$ is the number of spectral bands. Let $\mathbf{L}_{p}\in\mathbb{R}^{P^2\times 1}$ denote the co-registered LiDAR patch. The hidden dimension of the model is $D$. We propose IFGNet, an implicit spatial-frequency fusion framework for hyperspectral and LiDAR data based on Kolmogorov-Arnold Networks (KANs), LiDAR-guided implicit aggregation in spatial and frequency domains. 

\begin{figure*}[t]
    \centering
    \includegraphics[width=0.9\textwidth]{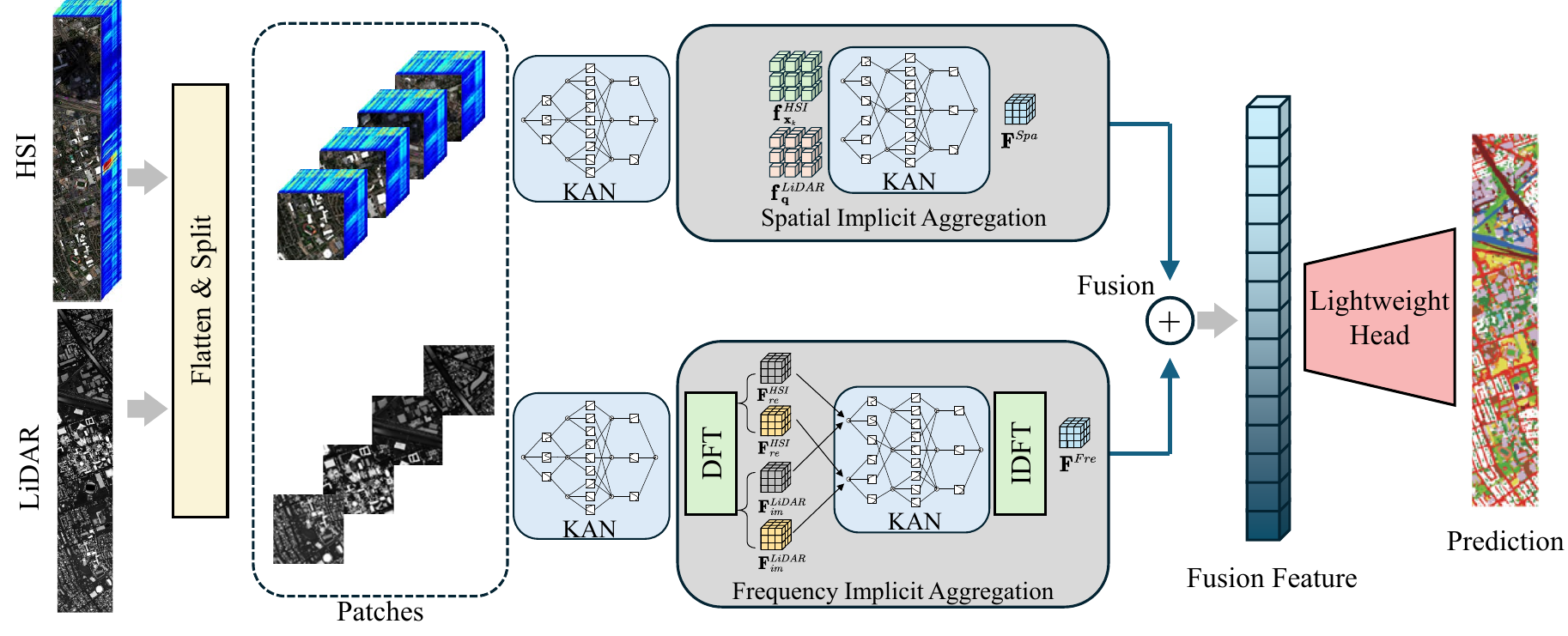}
    \caption{Overall architecture of IFGNet. The proposed framework performs geometry-aware
    and frequency-aware fusion of hyperspectral and LiDAR data via unified implicit representations.}
    \label{fig:ifgnet_framework}
\end{figure*}

As illustrated in Fig.~\ref{fig:ifgnet_framework}, the overall architecture of IFGNet consists of four core components: a KAN-based encoder for nonlinear feature learning, a spatial implicit aggregation module guided by LiDAR geometry, a frequency-domain implicit aggregation module for modeling global structures, and a lightweight head for classification. Specifically, hyperspectral and LiDAR features are first encoded into a shared latent space. Then, implicit aggregation is conducted in two complementary domains: (1) a spatial domain, where LiDAR-derived geometric information guides the fusion of spatial representations in HSI, and (2) a frequency domain, where FFT-based modeling captures global structures and long-range dependencies of the two modalities, and fuses both real and imagery components. The aggregated representations are finally classified by a lightweight head. Together, these components enable geometry-aware and frequency-aware fusion within a unified implicit representation framework.

\subsection{KAN-Based Encoder Layer}
\label{sec:encoder}

To enable flexible and continuous feature encoding for hyperspectral and LiDAR data, we employ Kolmogorov-Arnold Networks (KANs) as the encoder. By parameterizing nonlinear mappings with learnable spline functions, KANs provide expressive latent representations that are well-suited for subsequent implicit spatial-frequency fusion.

% Let $\mathbf{X}_p \in \mathbb{R}^{T \times P^2}$ denote an HSI patch with $T$ bands and $\mathbf{L}_p \in \mathbb{R}^{1 \times P^2}$ denote the aligned LiDAR patch. 

Each spectral vector of HSI and LiDAR value within the patch is independently mapped into a latent space of dimension $D$ using modality-specific KAN encoders:
\begin{equation}
\mathbf{F}^{HSI} = \Phi^{HSI}_{\text{KAN}}(\mathbf{X}_p), \quad
\mathbf{F}^{LiDAR} = \Phi^{LiDAR}_{\text{KAN}}(\mathbf{L}_p),
\end{equation}
where $\Phi^{HSI}_{\text{KAN}}(\cdot)$ and $\Phi^{LiDAR}_{\text{KAN}}(\cdot)$ denote KAN-based nonlinear mappings for hyperspectral and LiDAR data, respectively.
% The resulting latent feature maps satisfy
% \begin{equation}
% \mathbf{F}^{HSI}, \mathbf{F}^{LiDAR} \in \mathbb{R}^{D \times P^2}.
% \end{equation}
Compared to conventional multilayer perceptrons (MLPs), KAN-based encoders offer superior theoretical expressiveness for modeling the highly nonlinear spectral-geometric characteristics of HSI-LiDAR data. While MLPs rely on fixed activation functions and static weight matrices, KAN parameterizes each connection with a learnable B-spline function, enabling high-fidelity nonlinear fitting of subtle spectral variations as well as enhanced local sensitivity to geometric discontinuities in LiDAR elevation data due to the local support property of splines.

\subsection{Spatial Implicit Aggregation Unit}
\label{sec:SIAU}

The Spatial Implicit Aggregation Unit (SIAU) is designed to perform geometry-aware aggregation in the latent feature space. It fuses hyperspectral representations by exploiting LiDAR-derived geometric guidance through an implicit, continuous aggregation process implemented with KAN.

Given the hyperspectral and LiDAR latent features
\(
\mathbf{F}^{HSI}, \mathbf{F}^{LiDAR} \in \mathbb{R}^{ P^2 \times D},
\)
SIAU operates directly in the feature space and performs implicit aggregation for each query spatial coordinate $\mathbf{q}$ within the patch grid. For a query position $\mathbf{q}$, a local neighborhood $\mathcal{N}(\mathbf{q})$ is first identified. For each neighboring coordinate $\mathbf{x}_k \in \mathcal{N}(\mathbf{q})$, the corresponding hyperspectral latent feature $\mathbf{f}^{HSI}_{\mathbf{x}_k}$ is concatenated with the LiDAR latent feature at the query position $\mathbf{f}^{LiDAR}_{\mathbf{q}}$ and the relative coordinate offset $(\mathbf{q} - \mathbf{x}_k)$. The concatenated input is fed into a KAN-based implicit function to generate candidate features and associated aggregation weights:
\begin{equation}
\mathbf{v}^{(k)}_{\mathbf{q}} =
\Phi_{\text{KAN}}
\Big(
\big[
\mathbf{f}^{HSI}_{\mathbf{x}_k},
\mathbf{f}^{LiDAR}_{\mathbf{q}},
\mathbf{q} - \mathbf{x}_k
\big]
\Big),
\label{eq:sigsu_kan}
\end{equation}
where $\mathbf{v}^{(k)}_{\mathbf{q}} \in \mathbb{R}^{D+1}$, with the first $D$ channels corresponding to candidate latent features and the last channel encoding an unnormalized aggregation weight.
Following the implicit feature generation, the aggregation weights are normalized using a softmax function over the neighborhood, obtaining $\alpha^{(k)}_{\mathbf{q}}$.
% \begin{equation}
% \alpha^{(k)}_{\mathbf{q}} =
% \text{softmax}\big(\mathbf{v}^{(k)}_{\mathbf{q}}[D]\big),
% \end{equation}
% where $\mathbf{v}^{(k)}_{\mathbf{q}}[D]$ denotes the last channel of $\mathbf{v}^{(k)}_{\mathbf{q}}$.
The final spatially guided representation at position $\mathbf{q}$ is obtained via weighted summation of the candidate features:
\begin{equation}
\mathbf{F}^{Spa}_{\mathbf{q}} =
\sum_{\mathbf{x}_k \in \mathcal{N}(\mathbf{q})}
\alpha^{(k)}_{\mathbf{q}}
\cdot
\mathbf{v}^{(k)}_{\mathbf{q}}[1{:}D],
\label{eq:sigsu_out}
\end{equation}
where $\mathbf{v}^{(k)}_{\mathbf{q}}[1{:}D]$ denotes the feature channels.
From a functional perspective, SIAU instantiates a geometry-conditioned implicit operator that performs feature aggregation in the latent space. The use of KANs enables high-order nonlinear approximation over multivariate inputs, allowing for the modeling of complex interactions between hyperspectral radiometric patterns and LiDAR-defined geometric structures.

\subsection{Frequency-domain Implicit Aggregation}
\label{sec:figsu}

While spatial-domain implicit aggregation effectively captures local geometric structures, urban scenes also exhibit strong correlations between object geometry and frequency characteristics. Certain classes present distinct structural patterns that are more clearly separated in the frequency domain. To exploit this property, we further extend the implicit aggregation mechanism to the frequency domain. The same implicit aggregation operator $\text{SIA}(\cdot)$ is reused in the frequency domain, where the neighborhood is defined over the frequency components rather than the spatial coordinates.

% Given the encoded hyperspectral and LiDAR latent features $\mathbf{F}^{HSI}$ and $\mathbf{F}^{LiDAR}$, a two-dimensional Fourier transform is applied independently on each latent feature channel over the spatial dimensions:
% \begin{equation}
% \mathbf{F}^{HSI}_{real}, \mathbf{F}^{HSI}_{img} = \mathcal{F}(\mathbf{F}^{HSI}),
% \end{equation}
% \begin{equation}
% \mathbf{F}^{LiDAR}_{real}, \mathbf{F}^{LiDAR}_{img} = \mathcal{F}(\mathbf{F}^{LiDAR}),
% \end{equation}
% where $\mathbf{F}_{real}$ and $\mathbf{F}_{img}$ denote the real and imaginary components, respectively.
Given the encoded latent features $\mathbf{F}^{HSI}$ and $\mathbf{F}^{LiDAR}$, a 2D Discrete Fourier Transform (DFT) is applied to extract the frequency-domain components: $\mathbf{F}^m_{re}, \mathbf{F}^m_{im} = \mathcal{F}(\mathbf{F}^m)$ for $m \in \{HSI, LiDAR\}$, where $\mathbf{F}_{re}$ and $\mathbf{F}_{im}$ denote the real and imaginary components, respectively.

The implicit geometric aggregation operator is then applied to the frequency domain in a component-wise manner:
\begin{equation}
\mathbf{F}^{Fre}_{re} =
\text{SIA}(\mathbf{F}^{HSI}_{re}, \mathbf{F}^{LiDAR}_{re}),
\end{equation}
\begin{equation}
\mathbf{F}^{Fre}_{im} =
\text{SIA}(\mathbf{F}^{HSI}_{im}, \mathbf{F}^{LiDAR}_{im}).
\end{equation}
Finally, the frequency-guided representation is reconstructed through an inverse Discrete Fourier Transform (IDFT):
\begin{equation}
\mathbf{F}^{Fre} = \mathcal{F}^{-1}(\mathbf{F}^{Fre}_{re}, \mathbf{F}^{Fre}_{im}).
\end{equation}

% \subsection{Decoder and Classification Head}
% \label{sec:decoder}

% The Decoder integrates the spatial- and frequency-guided representations to generate a unified fusion feature. Specifically, an additive fusion strategy is employed to combine the spatial-domain and frequency-domain outputs, yielding the final fused representation $\hat{\mathbf{z}}$. The fused feature is subsequently fed into a lightweight classification head, where global average pooling (GAP) is first applied to aggregate spatial information, followed by a linear classifier and a softmax layer to predict land-cover categories:
% \begin{equation}
% \hat{\mathbf{y}} = \text{Softmax}\big( \mathbf{W} \cdot \text{GAP}(\hat{\mathbf{z}}) \big).
% \end{equation}
\subsection{Final Fusion and Classification Head}
\label{sec:decoder}
To leverage the complementary advantages of different domains, the spatially guided representation $\mathbf{F}^{Spa}$ and the frequency-guided representation $\mathbf{F}^{Fre}$ are integrated using an additive fusion strategy:\begin{equation}\hat{\mathbf{z}} = \mathbf{F}^{Spa} + \mathbf{F}^{Fre},\end{equation}where $\hat{\mathbf{z}}$ denotes the unified multi-modal fusion feature. This fused representation is subsequently fed into a lightweight classification head.

\section{Experiments}

We evaluate the proposed HSI-LiDAR fusion framework on two widely used HSI-LiDAR benchmarks,
namely the MUUFL Gulfport dataset and the Houston 2013 dataset, both providing spatially
co-registered hyperspectral imagery and LiDAR data. A patch-based pixel-wise classification
strategy is adopted, where for each labeled pixel, an HSI patch and its corresponding LiDAR patch
are extracted at the same spatial location. All samples are directly drawn from the original labeled regions, without additional data
augmentation.

For both datasets, data split and patch size strictly follow the experimental protocol in~\cite{zhang2025mamba}.
Specifically, a patch size of $9\times9$
is used for Houston 2013, while a smaller patch size of $5\times5$ is adopted for MUUFL. For dataset split, a class-balanced strategy is adopted by randomly selecting
150 labeled samples per class for training, with all remaining labeled samples used for testing the MUUFL dataset. For the Houston 2013 dataset, the standard training and testing splits 
are used, ensuring a fair and consistent comparison with existing methods.
All experiments are implemented using the PyTorch framework and conducted on a single NVIDIA
RTX 4090 GPU. Models are trained for 50 epochs using the Adam optimizer, and classification
performance is evaluated in terms of overall accuracy (OA), average accuracy (AA), and
Cohen’s Kappa coefficient ($\kappa$).

\subsection{Comparison Study}
We compare the proposed IFGNet with a broad range of representative HSI-LiDAR fusion methods, including CALC~\cite{lu2023coupled}, GLTNet~\cite{ding2022global}, M2FNet~\cite{SunMultiscale}, MS$^2$CANet~\cite{wang2024ms2canet}, S$^2$ENet~\cite{fang2021s2enet}, MFT~\cite{roy2023multimodal}, and ExViT~\cite{yao2023extended}. Table~\ref{tab:overall_results} reports the overall classification performance on the Houston 2013 and MUUFL datasets. The proposed IFGNet achieves the best performance across all metrics on both datasets, with OA reaching 99.37\% on Houston 2013 and 92.67\% on MUUFL. Significant improvements are observed for spectrally confusing urban areas, indicating that IFGNet effectively leverages LiDAR-derived geometric cues to enhance spatial consistency, while frequency-domain modeling further refines structural representation. The consistent gains on two benchmarks suggest strong robustness and generalization capability.

\begin{table}[t]
\centering
\caption{Overall classification performance (\%) on the Houston 2013 and MUUFL datasets.
The best results are in bold and the second best results are underlined.}
\label{tab:overall_results}
\setlength{\tabcolsep}{5.5pt}
\begin{tabular}{l c c c c c c}
\hline
Method 
& \multicolumn{3}{c}{Houston 2013} 
& \multicolumn{3}{c}{MUUFL} \\
\cline{2-7}
& OA & AA & Kappa & OA & AA & Kappa \\
\hline
CALC      & 88.12 & 90.02 & 87.12 & 90.94 & 91.60 & 88.08 \\
GLTNet    & 87.54 & 89.27 & 86.54 & 90.30 & 91.11 & 87.28 \\
M2FNet    & 89.46 & 90.93 & 88.58 & 85.38 & 89.91 & 81.22 \\
MS2CANet  & 92.19 & 93.09 & 91.58 & 90.85 & 91.41 & 88.05 \\
S$^2$ENet & 92.66 & 93.51 & 92.03 & 89.70 & 91.49 & 86.53 \\
MFT       & 91.25 & 92.23 & 90.51 & 86.28 & 86.67 & 82.17 \\
ExViT     & 92.31 & 93.31 & 91.67 & 87.92 & 90.32 & 84.36 \\
\hline
Ours      & \textbf{99.37} & \textbf{99.50} & \textbf{99.32}
          & \textbf{92.67} & \textbf{94.47} & \textbf{90.45} \\
\hline
\end{tabular}
\end{table}

\subsection{Ablation Study}
To investigate the contribution of each implicit aggregation component in IFGNet, we conduct an ablation study on the MUUFL dataset by selectively enabling the spatial-domain or frequency-domain implicit aggregation modules. As reported in Table~\ref{tab:ablation_spatial_frequency}, using only frequency-domain aggregation improves global structural discrimination but lacks fine-grained geometric guidance, while the spatial-only variant benefits from LiDAR-guided local aggregation yet misses global frequency-aware context. By jointly integrating both spatial and frequency implicit aggregation, IFGNet achieves the best overall performance across all evaluation metrics. These results demonstrate that the two implicit aggregation mechanisms are complementary and jointly essential for effective HSI-LiDAR fusion.
\begin{table}[t]
\centering
\caption{Ablation study of implicit aggregation modules on the MUUFL.}
\label{tab:ablation_spatial_frequency}
\begin{tabular}{lccc}
\hline
\textbf{Model Variant} & \textbf{OA (\%)} & \textbf{AA (\%)} & \textbf{Kappa (\%)} \\
\hline
Frequency-only & 91.00 & 91.66 & 88.28 \\
Spatial-only  & 91.94 & 92.25 & 84.26 \\
Spatial + Frequency (Full) & \textbf{92.67} & \textbf{94.47} & \textbf{90.45} \\
\hline
\end{tabular}
\end{table}

\subsection{Parameter Analysis}
To evaluate the effect of patch size, experiments were conducted on the MUUFL dataset, as summarized in Table~\ref{tab:patch_size_analysis_muufl}. A small patch size of 3 leads to inferior performance due to insufficient spatial–spectral context. Increasing the patch size to 5 significantly improves OA, AA, and Kappa, while a further increase to 7 results in marginal performance variations, likely caused by redundant background information. Therefore, a patch size of 5 is used as the default setting in subsequent experiment

\begin{table}[t]
\centering
\caption{Parameter analysis of patch size on the MUUFL dataset.}
\label{tab:patch_size_analysis_muufl}
\setlength{\tabcolsep}{6pt}
\begin{tabular}{c c c c}
\hline
\textbf{Patch Size} & \textbf{OA (\%)} & \textbf{AA (\%)} & \textbf{Kappa (\%)} \\
\hline
3  & 90.14 & 92.27 & 89.97 \\
\textbf{5}  & \textbf{92.67} & \textbf{94.47} & \textbf{90.45} \\
7  & 92.36 & 93.41 & 90.04 \\
\hline
\end{tabular}
\end{table}

\section{Conclusion}
In this paper, we propose an implicit spatial-frequency fusion framework for hyperspectral and LiDAR data classification based on Kolmogorov-Arnold Networks. By jointly modeling LiDAR-guided geometric aggregation in the spatial domain and implicit aggregation in the frequency domain, the proposed method effectively captures continuous spatial-spectral-geometric relationships. Extensive experiments on the Houston 2013 and MUUFL datasets demonstrate that the proposed approach consistently outperforms existing fusion methods. Ablation analyses further validate the complementary roles of spatial and frequency implicit aggregation. Future work will explore extending the proposed framework to large-scale scenes and other multimodal remote sensing tasks.

\bibliographystyle{IEEEtranN}
\bibliography{references}

\end{document}